\documentclass[11pt]{article}
\usepackage{acl2012}
\usepackage{natbib}
\renewcommand{\cite}[2][]{\citep[#1]{#2}}     
\renewcommand{\newcite}[2][]{\citet[#1]{#2}} 
 
\usepackage{amsmath,mathrsfs}
\usepackage{times}
\usepackage{url}
\usepackage{latexsym}
\usepackage{amssymb}
\usepackage{amsfonts}
\usepackage{mathtools}
\usepackage{multirow}
\usepackage{hhline}
\usepackage{graphicx}
\usepackage{verbatim} 
\usepackage{upgreek}
\usepackage{rotating}
\usepackage[usenames,dvipsnames]{color}
\usepackage{algorithm}
\usepackage[noend]{algpseudocode}
\usepackage{enumitem}
\usepackage{xspace}
\usepackage{tikz}
\usepackage{tikz-dependency}
\usepackage{appendix}
\usepackage[hidelinks]{hyperref}

\newcommand{\ourData}{Galactic Dependencies\xspace}
\newcommand{\N}{{\texttt{N}}}
\newcommand{\V}{{\texttt{V}}}
\newcommand{\defn}[1]{\textbf{#1}}
\newcommand{\RN}{R_{\texttt{N}}}
\newcommand{\RV}{R_{\texttt{V}}}
\renewcommand{\vec}[1]{{\boldsymbol{\mathbf{#1}}}}
\newcommand{\vtheta}{\vec{\theta}}
\newcommand{\vf}{\vec{f}}
\DeclareMathOperator*{\argmax}{argmax}
\DeclareMathOperator*{\mean}{mean}
\algnewcommand\algorithmicinput{\textbf{Input:}}
\algnewcommand\INPUT{\item[\algorithmicinput]}
\algnewcommand\algorithmicoutput{\textbf{Output:}}
\algnewcommand\OUTPUT{\item[\algorithmicoutput]}
\newcommand{\mc}[1]{\mathcal{#1}}

\title{{\bf The \ourData Treebanks: \\ Getting More Data by Synthesizing New
    Languages}}
\author{Dingquan Wang \and Jason Eisner  \\
        Department of Computer Science, Johns Hopkins University \\
        \texttt{\{wdd,eisner\}@jhu.edu}}
\date{\today{}}

\begin{document}

\maketitle

\begin{abstract}
  We release \ourData 1.0---a large set of synthetic languages not found on Earth, but annotated in Universal Dependencies format.  This new resource aims to provide training and development data for NLP methods that aim to adapt to unfamiliar languages.  Each synthetic treebank is produced from a real treebank by stochastically permuting the dependents of nouns and/or verbs to match the word order of other real languages. 
    We discuss the usefulness, realism, parsability, perplexity, and diversity of the synthetic languages.  
    As a simple demonstration of the use of \ourData, we consider single-source transfer, which attempts to parse a real target language using a parser trained on a ``nearby'' source language.  We find that including synthetic source languages somewhat increases the diversity of the source pool, which significantly improves results for most target languages.
\end{abstract}

\section{Motivation}\label{sec:intro}

Some potential NLP tasks have very sparse data by machine learning standards, as
each of the IID training examples is an {\em entire language}.  For instance:
\begin{itemize}[noitemsep]
\item typological classification of a language on various dimensions; 
\item adaptation of any existing NLP system to new, low-resource languages;
\item induction of a syntactic grammar from text;
\item discovery of a morphological lexicon from text;
\item other types of unsupervised discovery of linguistic structure.
\end{itemize}

Given a corpus or other data about a language, we might aim to predict whether it is an SVO language, or to learn to pick out its noun phrases.  For such problems, a single training or test example corresponds to an entire human language.

Unfortunately, we usually have only from 1 to 40 languages to work with.  In contrast, 
machine learning methods thrive on data, and 
recent AI successes have mainly been on tasks where one can train richly parameterized predictors on a huge set of IID (input, output) examples.  Even 7,000 training examples---one for each language or dialect on Earth---would be a small dataset by contemporary standards.  

As a result, it is challenging to develop systems that will discover structure in new languages in the same way that an image segmentation method, for example, will discover structure in new images.  The limited resources even make it challenging to develop methods that handle new languages by unsupervised, semi-supervised, or transfer learning.  Some such projects evaluate their methods on new sentences of the same languages that were used to develop the methods in the first place---which leaves one worried that the methods may be inadvertently tuned to the development languages and may not be able to discover correct structure in other languages.  Other projects take care to hold out languages for evaluation \cite{spitkovsky-2013-thesis,cotterell-peng-eisner-2015}, but then are left with only a few development languages on which to experiment with different unsupervised methods and their hyperparameters.

If we had many languages, then we could develop better unsupervised language learners.  Even better, we could treat linguistic structure discovery as a {\em supervised} learning problem.  That is, we could train a system to extract features from the surface of a language that are predictive of its deeper structure.  Principles \& Parameters theory \cite{chomsky_lectures_1981} conjectures that such features exist and that the juvenile human brain is adapted to extract them.

Our goal in this paper is to release a set of about 50,000 high-resource languages that could be used to train supervised learners, or to evaluate less-supervised learners during development.  These ``unearthly'' languages are intended to be at least {\em similar} to possible human languages.  As such, they provide useful additional training and development data that is slightly out of domain (reducing the variance of a system's learned parameters at the cost of introducing some bias). The initial release 
as described in this paper (version 1.0) is available at \url{https://github.com/gdtreebank/gdtreebank}.
We plan to augment this dataset in future work (\S\ref{sec:futurework}).

In addition to releasing thousands of treebanks, we provide scripts that can be used to ``translate'' other annotated resources into these synthetic languages.  E.g., given a corpus of English sentences labeled with sentiment, a researcher could reorder the words in each English sentence according to one of our English-based synthetic languages, thereby obtaining {\em labeled} sentences in the synthetic language.

\section{Related Work}\label{sec:relatedwork}

Synthetic data generation is a well-known trick for effectively training a large model on a small dataset. \newcite{abumostafa:1995} reviews early work that provided ``hints'' to a learning system in the form of virtual training examples.  While datasets have grown in recent years, so have models: e.g., neural networks have many parameters to train.
Thus, it is still common to create synthetic training examples---often by adding noise to real inputs or otherwise transforming them in ways that are expected to preserve their labels.  Domains where it is easy to exploit these invariances include image recognition \cite{simard2003best,NIPS2012_4824}, speech recognition \cite{jaitly2013vocal,Cui:2015:DAD:2824192.2824198}, information retrieval \cite{vilares-vilares-otero-2011}, and 
grammatical error correction \cite{rozovskaya-roth-2010}.

Synthetic datasets have also arisen recently for semantic tasks in natural language processing. bAbI is a dataset of facts, questions, and answers, generated by random simulation, for training machines to do simple logic \cite{weston-et-al-2016}. 
\newcite{DBLP:journals/corr/HermannKGEKSB15} generate reading comprehension questions and their answers, based on a large set of news-summarization pairs, for training machine readers. \newcite{serban2016generating} used RNNs to generate 30 million factoid questions about Freebase, with answers, for training question-answering systems. 
\newcite{wang2015building} obtain data to train semantic parsers in a new domain by first generating synthetic (utterance, logical form) pairs and then asking human annotators to paraphrase the synthetic utterances into more natural human language.

In speech recognition, morphology-based ``vocabulary expansion'' creates synthetic word forms \cite{rasooli-et-al-2014,varjokallio-klakow-2016}. 

Machine translation researchers have often tried to automatically preprocess parse trees of a source language to more closely resemble those of the target language, using either hand-crafted or automatically extracted rules (\citealp{dorr-et-al-2002,collins-koehn-kucerova-2005}, etc.; see review by \citealp{howlett-dras-2011}).

\section{Synthetic Language Generation}
\label{sec:synthetic_languages_generation}

A treebank is a corpus of parsed sentences of some language.  We propose to derive each synthetic treebank from some real treebank.  By manipulating the existing parse trees, we obtain a useful corpus for our synthetic language---a corpus that is already {\em tagged}, {\em parsed}, and {\em partitioned} into training/development/test sets.  
 Additional data in the synthetic language can be obtained, if desired, by automatically parsing additional real-language sentences and manipulating these trees in the same way.   

\begin{figure*}
\centering
\begin{dependency}[theme = simple]
  \tikzstyle{every node}=[font=\small]
   \begin{deptext}[column sep=.2em]
$*$\& \texttt{DET} \& \texttt{NOUN}   \& \texttt{PROPN}  \& \texttt{VERB} \& \texttt{VERB}  \& \texttt{DET}  \& \texttt{ADJ}   \& \texttt{NOUN}   \& \texttt{ADV}  \& \texttt{PUNCT}\\ 
\texttt{ROOT} \& \texttt{Every} \& \texttt{move} \& \texttt{Google} \& \texttt{makes} \& \texttt{brings} \& \texttt{this} \& \texttt{particular} \& \texttt{future} \& \texttt{closer} \& \texttt{.}\\
   \end{deptext}
   \depedge[arc angle=10]{3}{2}{det}
   \depedge{6}{3}{nsubj}  
   \depedge{5}{4}{nsubj} 
   \depedge{3}{5}{acl:rel}
   \depedge{1}{6}{root}
   \depedge{9}{7}{det}
   \depedge{9}{8}{amod}   
   \depedge{6}{9}{dobj}  
   \depedge{6}{10}{advmod} 
   \depedge{6}{11}{punct}
\end{dependency}
\begin{tabular}{|l|l|}
\hline
Language&Sentence\\\hline
English&Every move Google makes brings this particular future closer.\\
English[French/\N]&Every move Google makes brings this future \underline{particular} closer.\\
English[Hindi/\V]&Every move Google makes \underline{this particular future} \underline{closer} brings.\\
English[French/\N, Hindi/\V]&Every move Google makes \underline{this future \underline{particular}}\rule[-9pt]{0pt}{0pt} \underline{closer} brings.\\
\hline
\end{tabular}
\caption{\label{fig:dep} The original UD tree for a short English sentence, and its ``translations'' into three synthetic languages, which are obtained by manipulating the tree.  (Moved constituents are underlined.)  Each language has a different distribution over surface part-of-speech sequences.}

\end{figure*}

\subsection{Method}

We begin with the Universal Dependencies collection version 1.2 \cite{UNIVDEP-1.2,UNIVDEP-2016},\footnote{\url{http://universaldependencies.org}} or UD.  This provides manually edge-labeled dependency treebanks in 37 real languages, in a consistent style and format---the Universal Dependencies format.  An example appears in Figure~\ref{fig:dep}.

In this paper, we select a \defn{substrate} language $S$ represented in the UD treebanks, and systematically reorder the dependents of some nodes in the $S$ trees, to obtain trees of a \defn{synthetic} language $S'$.  

Specifically, we choose a \defn{superstrate} language $\RV$, and write $S' = S[\RV/\V]$ to denote a (projective) synthetic language obtained from $S$ by permuting the dependents of verbs (\V) to match the ordering statistics of the $\RV$ treebanks.  We can similarly permute the dependents of nouns (\N).\footnote{In practice, this means applying a single permutation model to permute the dependents of every word tagged as \texttt{NOUN} (common noun), \texttt{PROPN} (proper noun), or \texttt{PRON} (pronoun).}
This permutes about 93\% of $S$'s nodes (Table~\ref{tb:stats}), as UD treats adpositions and conjunctions as childless dependents.

For example, English[French/\N, Hindi/\V] is a synthetic language based on an English substrate, but which adopts subject-object-verb (SOV) word order from the Hindi superstrate and noun-adjective word order from the French superstrate (Figure~\ref{fig:dep}).  Note that it still uses English lexical items.

Our terms ``substrate'' and ``superstrate'' are borrowed from the terminology of creoles, although our synthetic languages are unlike naturally occurring creoles.  Our substitution notation $S' = S[\RN/\N,\RV/\V]$ is borrowed from the logic and programming languages communities.

\subsection{Discussion}\label{sec:proscons}

There may be more adventurous ways to manufacture synthetic languages (see \S\ref{sec:futurework} for some options).  However, we emphasize that our current method is designed to produce fairly {\em realistic} languages.  

First, we retain the immediate dominance structure and lexical items of the substrate trees, altering only their linear precedence relations.  Thus each sentence remains topically coherent; nouns continue to be distinguished by case according to their role in the clause structure; {\em wh}-words continue to c-command gaps; different verbs (e.g., transitive vs.\@ intransitive) continue to be associated with different subcategorization frames; and so on.  These important properties would not be captured by a simple context-free model of dependency trees, which is why we modify real sentences rather than generating new sentences from such a model.  In addition, our method obviously preserves the basic context-free properties, such as the fact that verbs typically subcategorize for one or two nominal arguments \cite{naseem_using_2010}.

Second, by drawing on real superstrate languages, we ensure that our synthetic languages use plausible word orders.  For example, if $\RV$ is a V2 language that favors SVO word order but also allows OVS, then $S'$ will match these proportions.  Similarly, $S'$ will place adverbs in reasonable positions with respect to the verb.  

We note, however, that our synthetic languages might violate some typological universals or typological tendencies.  For example, $\RV$ might prescribe head-initial verb orderings while $\RN$ prescribes head-final noun orderings, yielding an unusual language.  Worse, we could synthesize a language that uses free word order (from $\RV$) even though nouns (from $S$) are not marked for case.  Such languages are rare, presumably for the functionalist reason that sentences would be too ambiguous. One could automatically filter out such an implausible language $S'$, or downweight it, upon discovering that a parser for $S'$ 
was much less accurate on held-out data than a comparable parser for $S$.

We also note that our reordering method (\S\ref{sec:ordermodel}) does ignore some linguistic structure.  For example, we do not currently condition the order of the dependent subtrees on their heaviness or on the length of resulting dependencies, and thus we will not faithfully model phenomena like heavy-shift \cite{hawkins-1994,eisner-smith-2010-iwptbook}.  Nor will we model the relative order of adjectives.
  We also treat all verbs interchangeably, and thus use the same word orders---drawn from $\RV$---for both main clauses and embedded clauses.  This means that we will never produce a language like German (which uses V2 order in main clauses and SOV order in embedded clauses), even if $\RV=\text{German}$.  
All of these problems could be addressed by enriching the features that are described in the next section.

\section{Modeling Dependent Order}\label{sec:ordermodel}
\newcommand{\sPos}{\mc{X}}

Let $X$ be a part-of-speech tag, such as \texttt{Verb}.  
To produce a dependency tree in language $S' = S[R_X/X]$, we start with a projective dependency tree in language $S$.\footnote{Our method can only produce projective trees.  This is because it recursively generates a node's dependent subtrees, one at a time, in some chosen order.  Thus, to be safe, we only apply our method to trees that were originally projective.  See \S\ref{sec:futurework}.}  For each node $x$ in the tree that is tagged with $X$, we stochastically select a new ordering for its dependent nodes, including a position in this ordering for the head $x$ itself.  Thus, if node $x$ has $n-1$ dependents, then we must sample from a probability distribution over $n!$ orderings.

Our job in this section is to define this probability distribution.  Using $\pi = (\pi_1, \ldots, \pi_n)$ to denote an ordering of these $n$ nodes, we define a log-linear model over the possible values of $\pi$:
\begin{equation}\label{eqn:loglin}
  p_\vtheta(\pi \mid x) = \frac{1}{Z(x)} \exp \;\smashoperator[lr]{\sum_{1 \leq i < j \leq n}}\; \vtheta \cdot \vf(\pi,i,j)
\end{equation}
Here $Z(x)$ is the normalizing constant for node $x$.  $\vtheta$ is the parameter vector of the model.  $\vf$ extracts a sparse feature vector that describes the ordered pair of nodes $\pi_i, \pi_j$, where the ordering $\pi$ would place $\pi_i$ to the left of $\pi_j$.  

\subsection{Efficient sampling}

To sample exactly from the distribution $p_\vtheta$,\footnote{We could alternatively have used MCMC sampling.} we must explicitly compute all $n!$ unnormalized probabilities and their sum $Z(x)$.  

Fortunately, we can compute each unnormalized probability in just $O(1)$ amortized time, if we enumerate the $n!$ orderings $\pi$ using the Steinhaus-Johnson-Trotter algorithm \cite{sedgewick-1977}.  This enumeration sequence has the property that any two consecutive permutations $\pi, \pi'$ differ by only a single swap of some pair of adjacent nodes.  Thus their probabilities are closely related: 
  the sum in equation~\eqref{eqn:loglin} can be updated in $O(1)$ time by subtracting $\vtheta \cdot \vf(\pi,i,i+1)$ and adding $\vtheta \cdot \vf(\pi',i,i+1)$ for some $i$.  The other $O(n^2)$ summands are unchanged.

In addition, if $n \geq 8$, we avoid this computation by omitting the entire tree 
from our treebank; so we have at most $7! = 5040$ summands.

\subsection{Training parameters on a real language}

Our feature functions (\S\ref{sec:features}) are fixed over all languages.  They refer to the 17 node labels (POS tags) and 40 edge labels (dependency relations) that are used consistently throughout the UD treebanks.  

For each UD language $L$ and each POS tag $X$, we find parameters $\vtheta_X^L$ that globally maximize the unregularized log-likelihood:
\begin{equation}\label{eqn:obj}
\vtheta_X^L = \argmax_\vtheta \sum_x \log p_\vtheta(\pi_x \mid x)
\end{equation}
Here $x$ ranges over all nodes tagged with $X$ in the projective training trees of the $L$ treebank, omitting nodes with $n \geq 7$ for speed.

The expensive part of this computation is the gradient of $\log Z(x)$, which is an expected feature vector.  
To compute this expectation efficiently, we again take care to loop over the permutations in Steinhaus-Johnson-Trotter order.

A given language $L$ may not use all of the tags and relations.  Universal features that mention unused tags or relations do not affect \eqref{eqn:obj}, and their weights remain at 0 during training.

\subsection{$\!\!$Setting parameters of a synthetic language}

We use \eqref{eqn:loglin} to permute the $X$ nodes of substrate language $S$ into an order resembling superstrate language $R_X$.  In essence, this applies the $R_X$ ordering model to {\em out-of-domain data}, since the $X$ nodes may have rather different sets of dependents in the $S$ treebank than in the $R_X$ treebank.
We mitigate this issue in two ways.  

First, our ordering model \eqref{eqn:loglin} is designed to be more robust to transfer than, say, a Markov model.  The position of each node is influenced by all $n-1$ other nodes, not just by the two adjacent nodes.  As a result, the burden of explaining the ordering is distributed over more features, and we hope some of these features will transfer to $S$.  For example, suppose $R_X$ lacks adverbs and yet we wish to use $\vtheta_X^{R_X}$ to permute a sequence of $S$ that contains adverbs.  Even though the resulting order must disrupt some familiar non-adverb bigrams by inserting adverbs, other features---which consider non-adjacent tags---will still favor an $R_X$-like order for the non-adverbs.

Second, we actually sample the reordering from a distribution $p_\vtheta$ with an interpolated parameter vector $$\vtheta = \vtheta_X^{S'} = (1-\lambda)\vtheta_X^{R_X} + \lambda\vtheta_X^S,$$ where $\lambda=0.05$.  This gives a weighted product of experts, in which ties are weakly broken in favor of the substrate ordering.  (Ties arise when $R_X$ is unfamiliar with some tags that appear in $S$, e.g., adverb.)

\subsection{Feature Templates}\label{sec:features}

We write $t_i$ for the POS tag of node $\pi_i$, and $r_i$ for the dependency relation of $\pi_i$ to the head node.  If $\pi_i$ is itself the head, then necessarily $t_i=X$,\footnote{Recall that for each head POS $X$ of language $L$, we learn a {\em separate} ordering model with parameter vector $\vtheta_X^L$.} and we specially define $r_i=\texttt{head}$.  

In our feature vector $\vf(\pi,i,j)$, the features with the following names have value 1, while all others have value 0:

\begin{itemize}
  \item \texttt{L.$t_i$.$r_i$} and \texttt{L.$t_i$} and \texttt{L.$r_i$}, provided that $r_j=\texttt{head}$.  For example, \texttt{L.ADJ} will fire on each \texttt{ADJ} node to the left of the head.
  \item \texttt{L.$t_i$.$r_i$.$t_j$.$r_j$} and \texttt{L.$t_i$.$t_j$} and \texttt{L.$r_i$.$r_j$}, provided that $r_i\neq\texttt{head},
 r_j\neq\texttt{head}$.  These features detect the relative order of two siblings. 
  \item $d.t_i.r_i.t_j.r_j$, $d.t_i.t_j$, and $d.r_i.r_j$, where $d$ is \texttt{l} (left), \texttt{m} (middle), or \texttt{r} (right) according to whether the head position $h$ satisfies $i < j < h$, $i < h < j$, or $h < i < j$.  For example, \texttt{l.nsubj.dobj} will fire on SOV clauses.  This is a specialization of the previous feature, and is skipped if $i=h$ or $j=h$.
  \item \texttt{A.$t_i$.$r_i$.$t_j$.$r_j$} and \texttt{A.$t_i$.$t_j$} and \texttt{A.$r_i$.$r_j$}, provided that $j=i+1$.  These ``bigram features'' detect two adjacent nodes.  For this feature and the next one, we extend the summation in \eqref{eqn:loglin} to allow $0 \leq i < j \leq n+1$, taking $t_0=r_0=\texttt{BOS}$ (``beginning of sequence'') and $t_{n+1}=r_{n+1}=\texttt{EOS}$ (``end of sequence'').  Thus, a bigram feature such as \texttt{A.DET.EOS} would fire on \texttt{DET} when it falls at the end of the sequence.
  \item \texttt{H.$t_i$.$r_i$.$t_{i+1}$.$r_{i+1}$.\ldots.$t_j$.$r_j$}, provided that $i+2 \leq j \leq i+4$.  Among features of this form, we keep only the 10\% that fire most frequently in the training data.  These ``higher-order $k$-gram'' features memorize sequences of lengths 3 to 5 that are common in the language.  
\end{itemize}
  
\noindent Notice that for each non-\texttt{H} feature that mentions both tags $t$
and relations $r$, we also defined two \defn{backoff} features,
omitting the $t$ fields or $r$ fields respectively.

Using the example from Figure \ref{fig:dep}, for subtree
\begin{center}
\begin{dependency}[theme = simple]
  \tikzstyle{every node}=[font=\small]
   \begin{deptext}[column sep=.2em]
\texttt{DET} \& \texttt{ADJ} \& \texttt{NOUN}\\ 
\texttt{this} \& \texttt{particular} \& \texttt{future}\\
   \end{deptext}
   \depedge{3}{1}{\normalsize det}  
   \depedge{3}{2}{\normalsize amod}  
\end{dependency}
\end{center}
the features that fire are

\begin{table}[!ht]
  \centering
\begin{small}
\begin{tabular}{|l|l|}
\hline
Template&Features\\
\hline
\texttt{L.$t_i$.$r_i$}&\texttt{L.DET.det}, \texttt{L.ADJ.amod}\\\hline 
\texttt{L.$t_i$.$r_i$.$t_j$.$r_j$}&\texttt{L.DET.det.ADJ.amod}\\\hline
\texttt{$d$.$t_i$.$r_i$.$t_j$.$r_j$}&\texttt{l.DET.det.ADJ.amod}\\ \hline
\texttt{A.$t_1$.$r_1$.$t_2$.$r_2$}&\parbox[t]{2cm}{\texttt{A.BOS.BOS.DET.det},\\
\texttt{A.DET.det.ADJ.amod},\\\texttt{A.ADJ.amod.NOUN.head},\\\texttt{A.NOUN.head.EOS.EOS}\vspace{1mm}}\\\hline
\end{tabular}
\end{small}
\end{table}

\noindent
plus backoff features and \texttt{H} features (not shown).

\section{The Resource}\label{sec:resource}

In \ourData v1.0, or GD, we release real and synthetic treebanks based on UD v1.2.  Each synthetic treebank is a modified work that is freely licensed under the same CC or GPL license as its substrate treebank.  We provide {\em all} languages of the form $S$, $S[\RV/\N]$, $S[\RN/\V]$, and $S[\RN/\N,\RV/\V]$, where the substrate $S$ and the superstrates $\RN$ and $\RV$ each range over the 37 available languages. ($\RN=S$ or $\RV=S$ gives ``self-permutation'').
  This yields $37 \times 38 \times 38 = 53,428$ languages in total.

Each language is provided as a directory of 3 files: training, development, and test treebanks.  The directories are systematically named: for example, English[French/\N, Hindi/\V] can be found in directory \texttt{en$\sim$fr@N$\sim$hi@V}.

\begin{table}[t]
  \centering
\begin{tabular}{|c|c|c|}
\hline
Train&Dev&Test\\
\hline
\parbox[t]{2cm}{\raggedright cs, es, fr, hi, de, it, la\_itt, no, ar, pt}&\parbox[t]{2.1cm}{\raggedright en, nl, da, fi, got, grc, et, la\_proiel, grc\_proiel, bg}&\parbox[t]{2.4cm}{\raggedright la, hr, ga, he, hu, fa, ta, cu, el, ro, sl, ja\_ktc, sv, fi\_ftb, id, eu, pl}\\
\hline
\end{tabular}
\caption{\label{tb:split}The 37 real UD languages.  Following the usual setting of rich-to-poor transfer, we take the 10 largest non-English languages (left column) as our pool of real source languages, which we can combine to synthesize new languages.  The remaining languages are our low-resource target languages.  We randomly hold out 17 non-English languages (right column) as the test languages for our final result table.  During development, we studied and graphed performance on the remaining 10 languages (middle column)---including English for interpretability.}
\end{table}

Our treebanks provide alignment information, to facilitate error analysis as well as work on machine translation.  Each word in a synthetic sentence is annotated with its original position in the substrate sentence.  Thus, all synthetic treebanks derived from the same substrate treebank are node-to-node aligned to the substrate treebank and hence to one another.

In addition to the generated data, we also provide the parameters $\theta_X^L$ of our ordering models; code for training new ordering models; and code for producing new synthetic trees and synthetic languages.  Our code should produce reproducible results across platforms, thanks to Java's portability and our standard random number seed of 0.

\begin{table}[t]
  \small
\setlength\tabcolsep{2pt}
\centering
\begin{tabular}{|c|r@{\,/\,}l|r@{\,/\,}l|c|r@{\,/\,}l|c|}
\hline
lang&\multicolumn{2}{c|}{sents}&\multicolumn{2}{c|}{tokens}&$T$ &\multicolumn{2}{c|}{UAS } &$R$\\ \hline
ar&4K&6K&119K&226K&85\%&72\%&69\%&0.37\\
cs&5K&7K&687K&1173K&94\%&81\%&78\%&0.38\\
de&9K&14K&136K&270K&94\%&84\%&80\%&0.47\\
es&10K&14K&211K&382K&94\%&85\%&82\%&0.32\\
fr&8K&15K&154K&356K&95\%&86\%&84\%&0.27\\
hi&9K&13K&160K&281K&96\%&82\%&82\%&0.20\\
it&9K&12K&144K&249K&95\%&87\%&84\%&0.30\\
la\_itt&7K&15K&87K&247K&90\%&66\%&58\%&0.72\\
no&11K&16K&135K&245K&93\%&82\%&79\%&0.31\\
pt&5K&9K&87K&202K&96\%&86\%&84\%&0.32\\
\hline
\end{tabular}
\caption{\label{tb:stats} Some statistics on the 10 real training languages. 
When two numbers are separated by ``/'', the second represents the full UD treebank, and the first comes from our GD version, which discards non-projective trees and high-fanout trees ($n \geq 8$).  UAS is the language's \defn{parsability}:   the unlabeled attachment score on its dev sentences after training on its train sentences.  $T$ is the percentage of GD tokens that are touched by reordering (namely $\N$, $\V$, and their dependents).
 $R \in [0,1]$ measures the \defn{freeness} of the language's word order, as the conditional cross-entropy of our trained ordering model $p_\vtheta$ relative to that of a uniform distribution: $R = \frac{H(\tilde{p},p_\vtheta)}{H(\tilde{p},p_{\text{unif}})} = \frac{\mean_x[-\log_2 p_\vtheta(\pi^*(x) \mid x)]}{\mean_x [-\log_2 1/n(x)!]}$, where $x$ ranges over all $\N$ and $\V$ tokens in the dev sentences, $n(x)$ is 1 + the number of dependents of $x$, and $\pi^*(x)$ is the observed ordering at $x$.}
\end{table}

\section{Exploratory Data Analysis}\label{sec:eda}

How do the synthetic languages compare to the real ones?  For analysis and experimentation, we partition the real UD languages into train/dev/test (Table \ref{tb:split}).  (This is orthogonal to the train/dev/test split of each language's treebank.)  Table \ref{tb:stats} shows some properties of the real training languages.

In this section and the next, we use the Yara parser \cite{rasooli-tetrault-2015}, a fast arc-eager transition-based projective dependency parser, with beam size of 8.  We train only \defn{delexicalized} parsers, whose input is the sequence of POS tags.  Parsing accuracy is evaluated by the unlabeled attachment score (UAS), that is, the fraction of word tokens in held-out (dev) data that are assigned their correct parent.  For language modeling, we train simple trigram backoff language models with add-1 smoothing, and we measure predictive accuracy as the perplexity of held-out (dev) data.

\begin{figure}[t]
\centering  
\includegraphics[width=8.0cm, height=4.0cm]{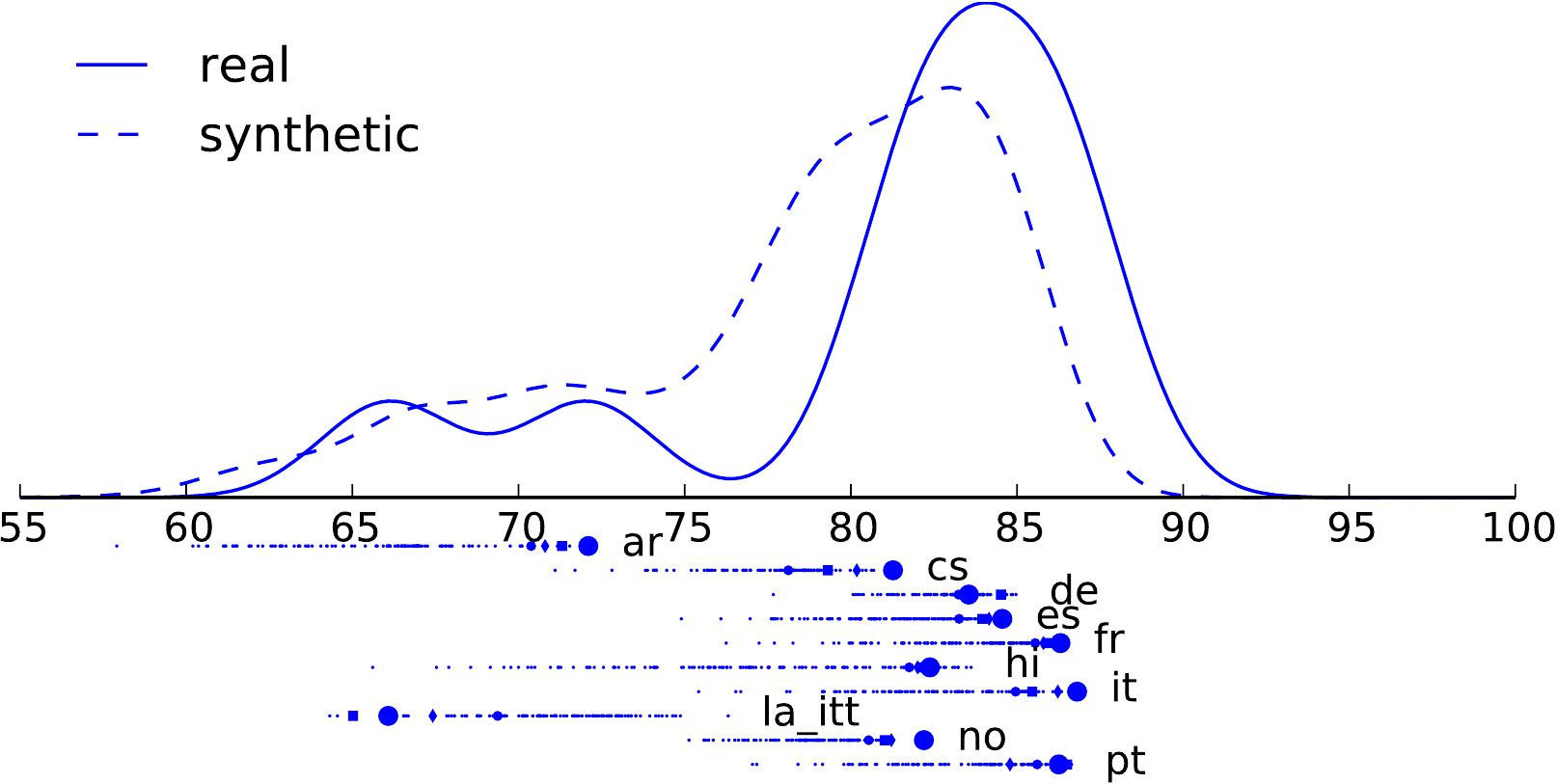}
\caption[]{\label{fig:lang_parse_density} Parsability of real versus synthetic languages (defined as in Table~\ref{tb:stats}).  The upper graphs are kernel density estimates.  Each lower graph is a 1-dimensional scatterplot, showing the parsability of some real language $S$ (large dot) and all its permuted versions, including the ``self-permuted'' languages $S[S/\N]$ (diamond), $S[S/\V]$ (square), and $S[S/\N,S/\V]$ (medium dot).}
\end{figure}

\begin{figure}[t]
\centering  
\includegraphics[width=8.0cm, height=4.5cm]{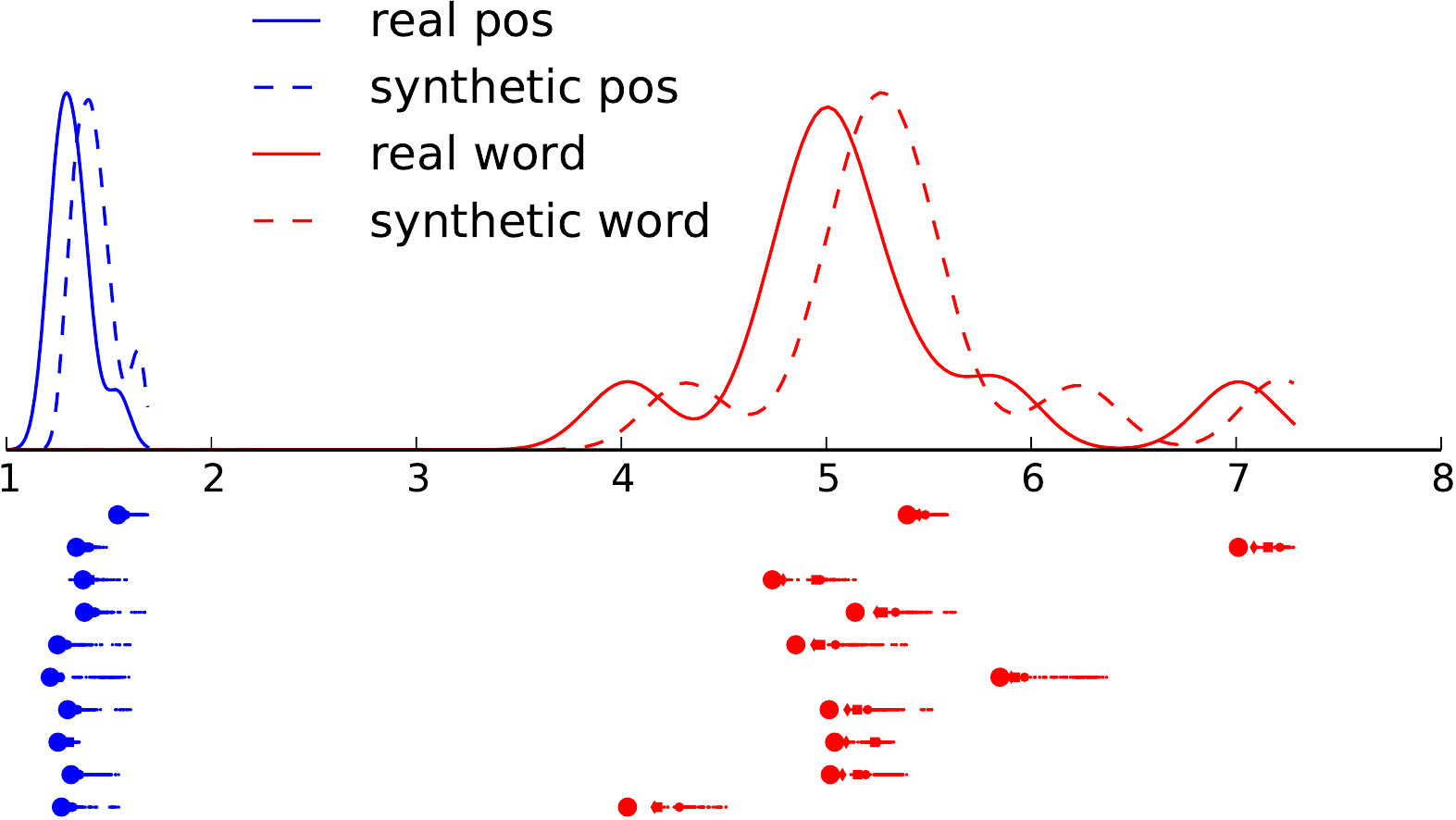}
\caption[]{\label{fig:lang_lm_density} Perplexity of the POS tag sequence, as well as the word sequence, of real versus synthetic languages.  Words with count $< 10$ are mapped to an OOV symbol.}
\end{figure}

Figures \ref{fig:lang_parse_density}--\ref{fig:lang_lm_density} show how the parsability and perplexity of a real training language usually get worse when we permute it.  
We could have discarded low-parsability synthetic languages, on the functionalist grounds that they would be unlikely to survive as natural languages anywhere in the galaxy.  However, the curves in these figures show that most synthetic languages have parsability and perplexity within the plausible range of natural languages, so we elected to simply keep all of them in our collection.

An interesting exception in Figure \ref{fig:lang_parse_density} is Latin (la\_itt), whose poor parsability---at least by a delexicalized parser that does not look at word endings---may be due to its especially free word order (Table~\ref{tb:stats}).  When we impose another language's more consistent word order on Latin, it becomes more parsable.  Elsewhere, permutation generally hurts, perhaps because a real language's word order is globally optimized to enhance parsability.  It even hurts slightly when we randomly ``self-permute'' $S$ trees to use other word orders that are common in $S$ itself!  Presumably this is because 
 the authors of
 the original $S$ sentences chose, or were required, to order each constituent in a way that would enhance its parsability in context: see the last paragraph of \S\ref{sec:proscons}.

\begin{figure}[t]
\centering  
\includegraphics[width=6.0cm, height=6.0cm]{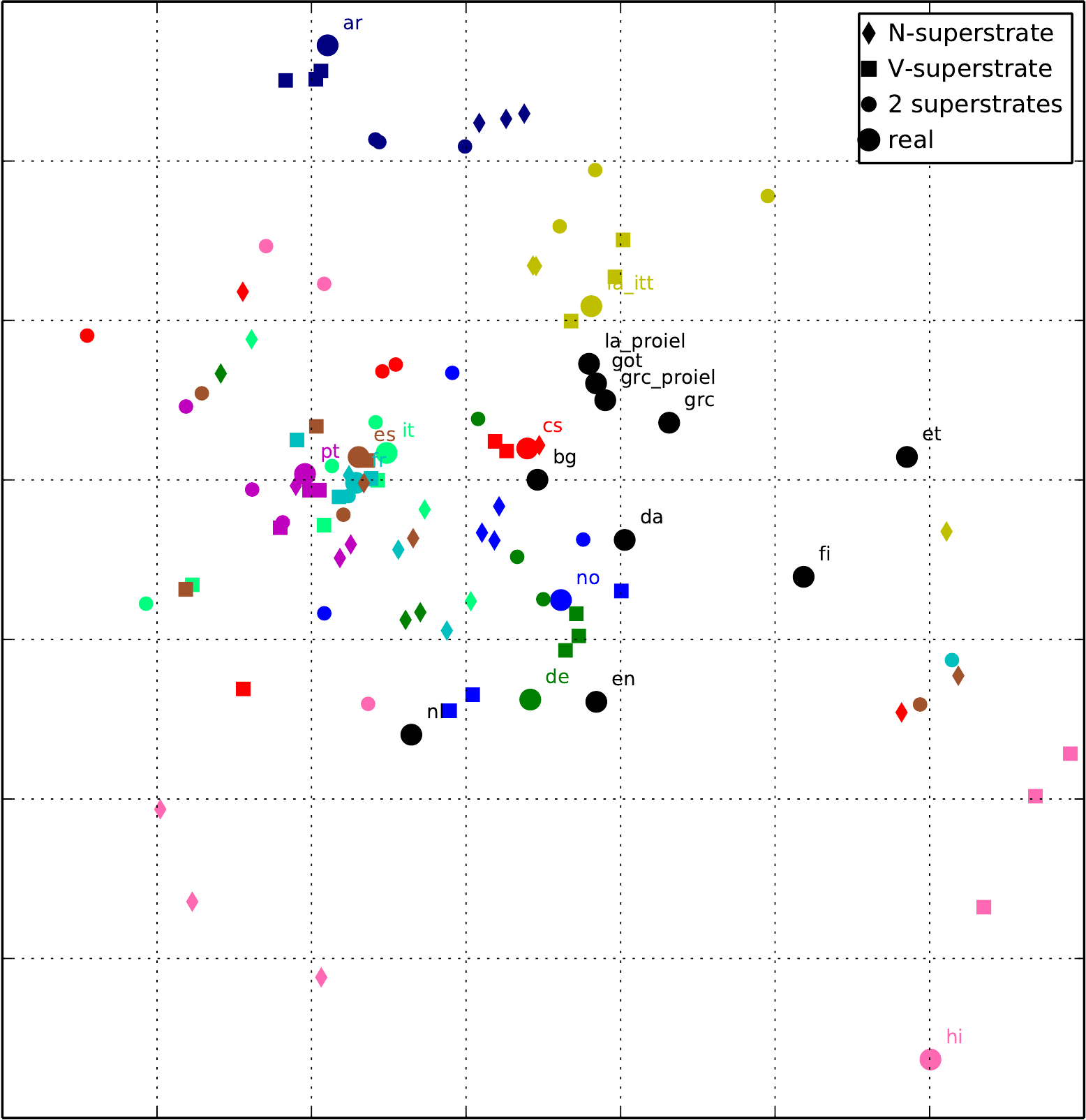}
\caption[]{\label{fig:distance_parse}  Each point represents a language.  The color of a synthetic language is the same as its substrate language.  Dev languages are shown in black.  This 2-dimensional embedding was constructed using metric multidimensional scaling \cite{BorgGroenen2005} on a symmetrized version of our dissimilarity matrix (which is not itself a metric).  The embedded distances are reasonably faithful to the symmetrized dissimilarities: metric MDS achieves a low value of 0.20 on its ``stress'' objective, and we find that Kendall's tau = 0.76, meaning that if one pair of languages is displayed as farther apart than another, then in over $7/8$ of cases, that pair is in fact more dissimilar.  Among the real languages, note the clustering of Italic languages (pt, es, fr, it), Germanic languages (de, no, en, nl, da), Slavic languages (cs, bg), and Uralic languages (et, fi).  Outliers are Arabic (ar), the only Afroasiatic language here, and Hindi (hi), the only SOV language, whose permutations are less outr\'{e} than it is.}
\vspace{-1pt}
\end{figure}

Synthesizing languages is a balancing act.  The synthetic languages are not useful if all of them are too conservatively close to their real sources to add diversity---or too radically different to belong in the galaxy of natural languages.  Fortunately, we are at neither extreme.  Figure~\ref{fig:distance_parse} visualizes a small sample of 110 languages from our collection.\footnote{For each of the 10 real training languages, we sampled 9 synthetic languages: 3 $\N$-permuted, 3 $\V$-permuted and 3 $\{\N,\V\}$-permuted. We also included all 10 training + 10 dev languages.}  For each ordered pair of languages $(S,T)$, we defined the dissimilarity $d(S,T)$ as the {\em decrease} in UAS when we parse the dev data of $T$ using a parser trained on $S$ instead of one trained on $T$.  Small dissimilarity (i.e., good parsing transfer) translates to small distance in the figure.  
The figure shows that the permutations of a substrate language (which share its color) can be radically different from it, as we already saw above.  Some may be unnatural, but others are similar to other real languages, including held-out dev languages.  Thus Dutch (nl) and Estonian (et) have close synthetic neighbors within this small sample, although they have no close real neighbors.

\section{An Experiment}

We now {\em illustrate} the use of GD by studying how expanding the set of available treebanks can improve a simple NLP method, related to Figure~\ref{fig:distance_parse}.

\subsection{Single-source transfer}

Dependency parsing of low-resource languages has been intensively studied for years.  A simple method is called ``single-source transfer'': parsing a target language $T$ with a parser that was trained on a source language $S$, where the two languages are syntactically similar.  Such single-source transfer parsers \cite{ganchev2010posterior,mcdonald2011multi,ma2014unsupervised,Guo:2015:clnndep,duong2015cross,rasoolidensity} are not state-of-the-art, but they have shown substantial improvements over fully unsupervised grammar induction systems \cite{klein_corpus-based_2004,smith_annealing_2006,spitkovsky-alshawi-jurafsky-2013}.

It is permitted for $S$ and $T$ to have different vocabularies.  The $S$ parser can nonetheless parse $T$ (as in Figure~\ref{fig:distance_parse})---provided that it is a ``delexicalized'' parser that only cares about the POS tags of the input words.  In this case, we require only that the target sentences have already been POS tagged using the same tagset as $S$: in our case, the UD tagset.

\subsection{Experimental Setup}

We evaluate single-source transfer when the pool of $m$ source languages consists of $n$ real UD languages, plus $m-n$ synthetic GD languages derived by ``remixing'' just these real languages.\footnote{The $m-n$ GD treebanks are comparatively impoverished because---in the current GD release---they include only projective sentences (Table~\ref{tb:stats}).  The $n$ UD treebanks are unfiltered.}  We try various values of $n$ and $m$, where $n$ can be as large as 10 (training languages from Table~\ref{tb:split}) and $m$ can be as large as $n \times (n+1) \times (n+1) \leq 1210$ (see \S\ref{sec:resource}).

Given a {\em real} target language $T$ from outside the pool, we {\em select} a single source language $S$ from the pool, and try to parse UD sentences of $T$ with a parser trained on $S$.  We evaluate the results on $T$ by measuring the unlabeled attachment score (UAS), that is, the fraction of word tokens that were assigned their correct parent.  In these experiments (unlike those of \S\ref{sec:eda}), we always evaluate fairly on $T$'s {\em full} dev or test set from UD---not just the sentences we kept for its GD version (cf.\@ Table~\ref{tb:stats}).\footnote{The Yara parser can only produce projective parses.  It attempts to parse all test sentences of $T$ projectively, but sadly ignores non-projective training sentences of $S$ (as can occur for real $S$).}

The hope is that a large pool will contain at least one language---real or synthetic---that is ``close'' to $T$.  We have two ways of trying to select a source $S$ with this property:

\defn{Supervised selection} selects the $S$ whose parser achieves the highest UAS on 100 training sentences of language $T$.  This requires 100 good trees for $T$, which could be obtained with a modest investment---a single annotator attempting to follow the UD annotation standards in a consistent way on 100 sentences of $T$, without writing out formal $T$-specific guidelines.  
(There is no guarantee that selecting a parser on {\em training} data will choose well for the {\em test} sentences of $T$.  We are using a small amount of data to select among {\em many} dubious parsers, many of which achieve similar results on the training sentences of $T$.  Furthermore, in the UD treebanks, the test sentences of $T$ are sometimes drawn from a different distribution than the training sentences.)

\defn{Unsupervised selection} selects the $S$ whose training sentences had the best ``coverage'' of the POS tag sequences in the actual data from $T$ that we aim to parse.  More precisely, we choose the $S$ that maximizes $p_S($tag sequences from $T)$---in other words, the maximum-likelihood $S$---where $p_S$ is our trigram language model for the tag sequences of $S$.
This approach is loosely inspired by \newcite{sogaard2011data}.

\subsection{Results}

\begin{figure}[!t]
\centering  
\includegraphics[width=8.0cm, height=7cm]{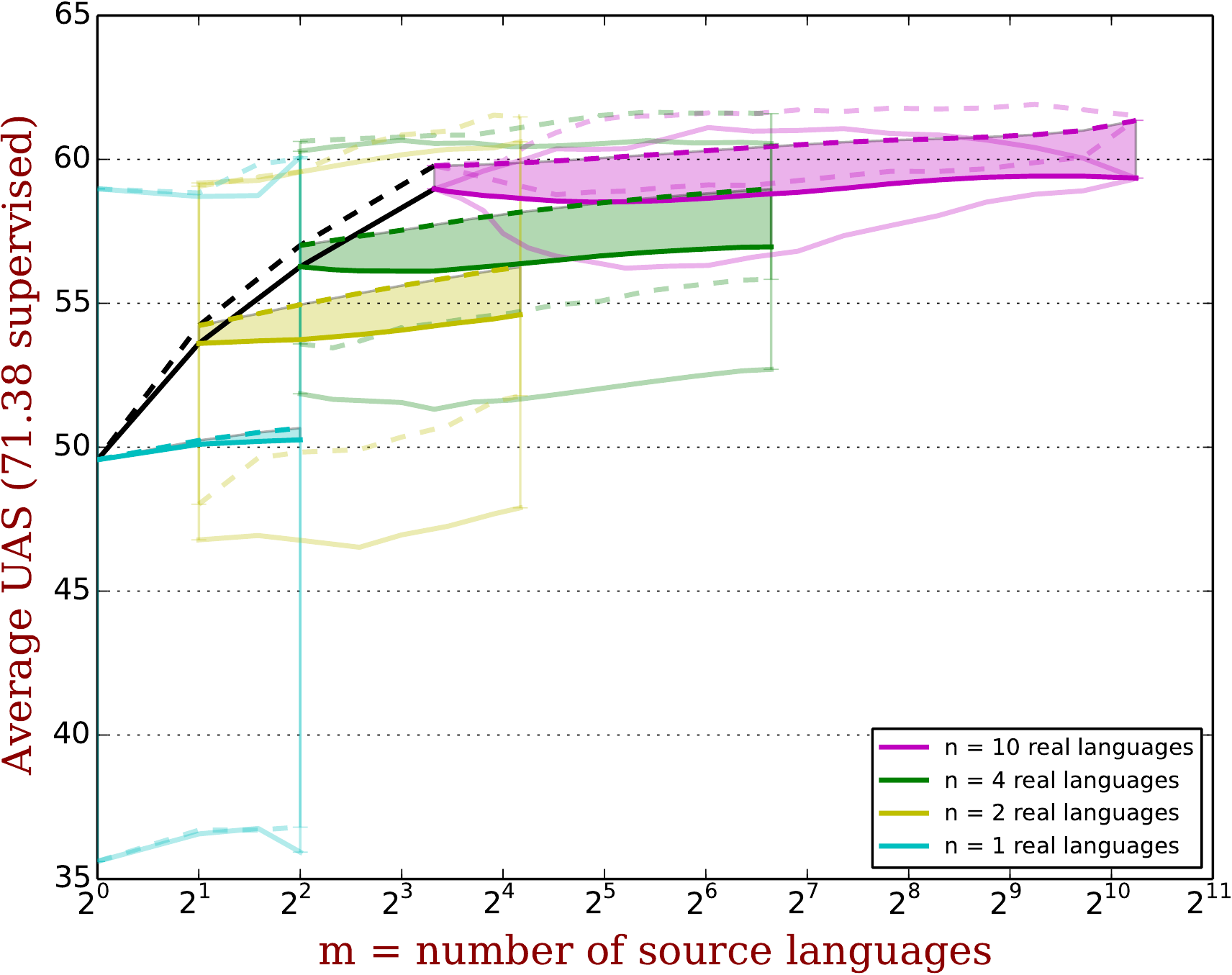}
\caption[]{\label{fig:kite}Comprehensive results for single-source transfer from a pool of $m$ languages (the horizontal axis) synthesized from $n$ real languages.  For each color $1,2,\ldots,n$, the upper dashed line shows the UAS achieved by supervised selection; the lower solid line shows unsupervised selection; and the shaded area highlights the difference. The black dashed and solid lines connect the points where $m=n$, showing how rapidly UAS increases with $n$ when only real languages are used. 
  \\ \hspace*{2ex}Each point is the {\em mean} dev UAS over 10,000 experiments.  We use paler lines in the same color and style to show the considerable {\em variance} of these UAS scores.  These essentially delimit the interdecile range from the 10th to the 90th percentile of UAS score.  However, if the plot shows a mean of 57, an interdecile range from 53 to 61 actually means that the middle 80\% of experiments were within $\pm 4$ percentage points of the mean UAS {\em for their target language}.  (In other words, before computing this range, we adjust each UAS score for target $T$ by subtracting the mean UAS from the experiments with target $T$, and adding back the mean UAS from all 10,000 experiments (e.g., 57).)
  \\ \hspace*{2ex}Notice that on the $n=10$ curve, there is no variation among experiments either at the minimum $m$ (where the pool always consists of all 10 real languages) or at the maximum $m$ (where the pool always consists of all 1210 galactic languages).
}
\end{figure}

Our most complete visualization is Figure~\ref{fig:kite}, which we like to call the ``kite graph'' for its appearance. We plot the UAS on the development treebank of $T$ as a function of $n$, $m$, and the selection method.  As Appendix~\ref{app:kite} details, each point on this graph is actually an average over 10,000 experiments that make random choices of $T$ (from the UD development languages), the $n$ real languages (from the UD training languages), and the $m-n$ synthetic languages (from the GD languages derived from the $n$ real languages).   
We see from the black lines that increasing the number of real languages $n$ is most beneficial.  But crucially, when $n$ is fixed in practice, gradually increasing $m$ by remixing the real languages does lead to meaningful improvements.  This is true for both selection methods.  Supervised selection is markedly better than unsupervised.

\begin{figure}[t]
\centering  
\includegraphics[width=8.0cm, height=7cm]{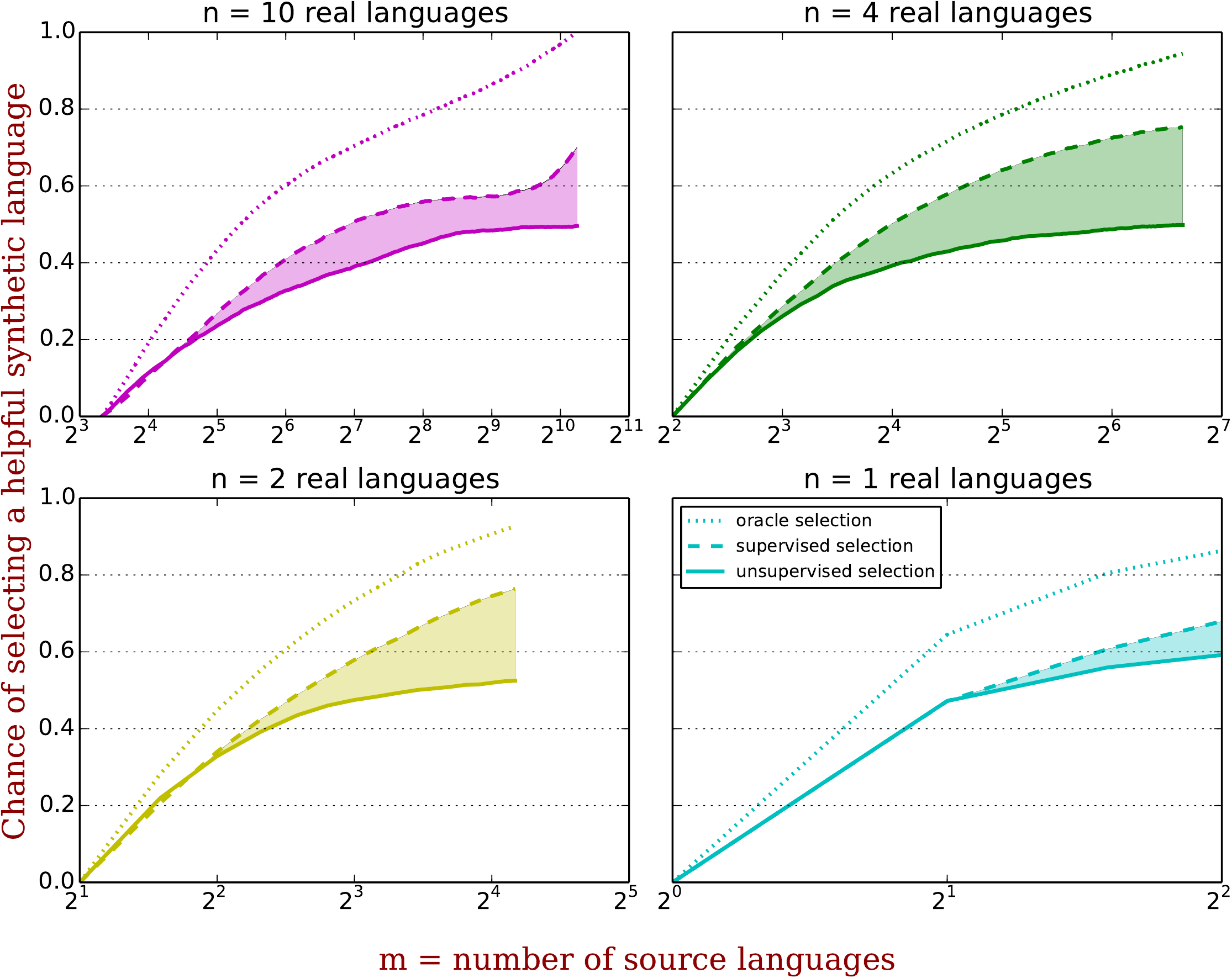}
\caption{\label{fig:selection} Chance that selecting a source from $m$ languages achieves strictly better dev UAS than just selecting from the $n$ real languages.}
\end{figure}

The ``selection graph'' in Figure~\ref{fig:selection} visualizes the same experiments in a different way.  Here we ask about the fraction of experiments in which using the full pool of $m$ source languages was strictly better than using only the $n$ real languages.  We find that when $m$ has increased to its maximum, the full pool nearly always contains a synthetic source language that gets better results than anything in the real pool.  After all, our generation of ``random'' languages is a scattershot attempt to hit the target: the more languages we generate, the higher our chances of coming close.  However, our selection methods only manage to {\em pick} a better language in about 60\% of those experiments.

\begin{figure}[!t]
\centering  
\includegraphics[width=8.0cm, height=6cm]{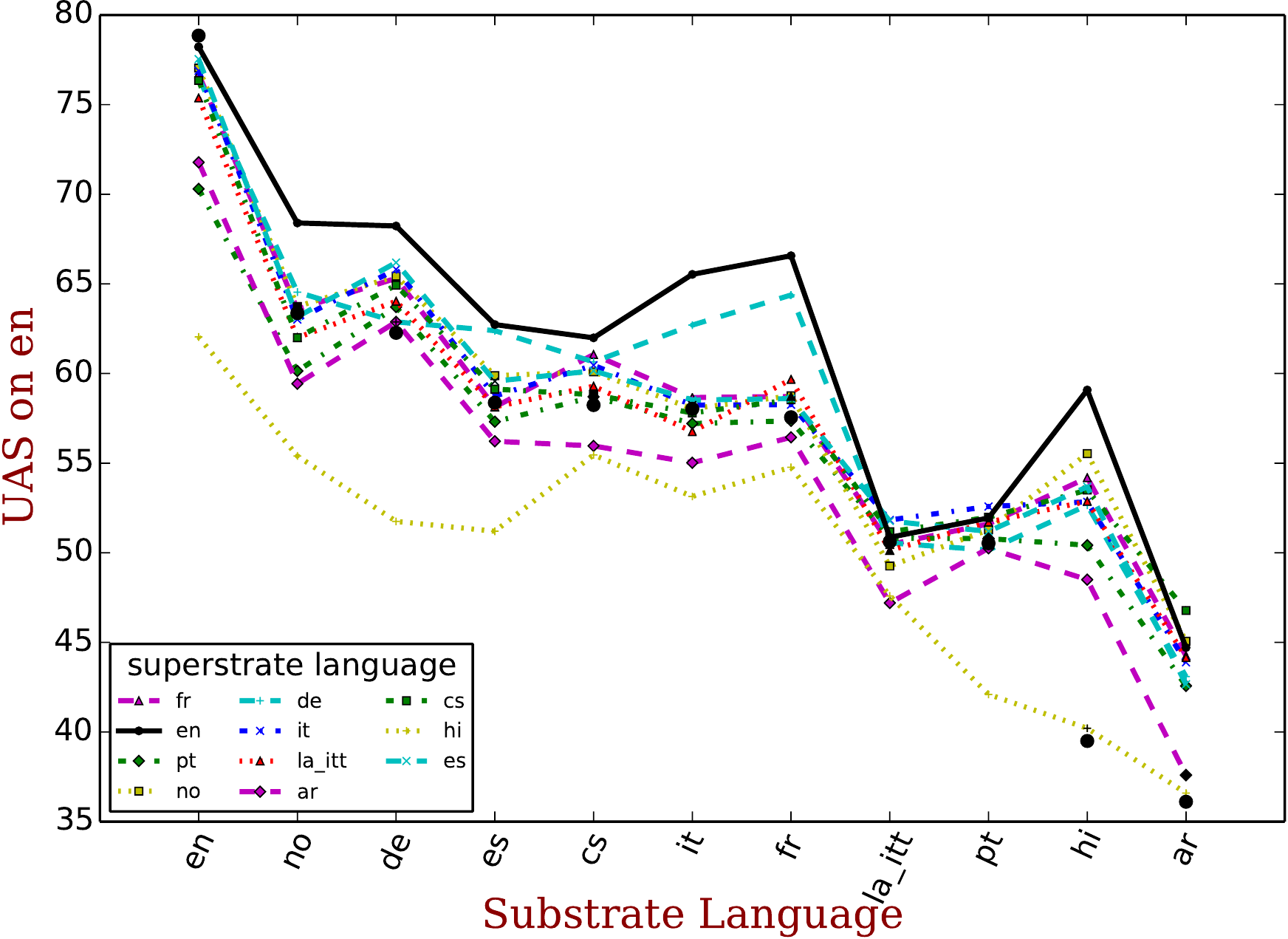}
\caption{\label{fig:tgt}UAS performance of different source parsers when applied to English development sentences.  The $x$ axis shows the 10 real training languages $S$, in decreasing order of their UAS performance (plotted as large black dots).  
  For each superstrate $R$, we plot a curve showing---for each substrate $S$---the best UAS of the languages $S[R/\N], S[R/\V]$ and $S[R/\N,R/\V]$.  The points where $R=S$ are specially colored in black; these are instances of {\em self-permutation} (\S\ref{sec:resource}).  We also add ``cheating results'' where English itself is used as the substrate (left column) and/or the superstrate (solid black line at top).  Thus, the large black dot at the upper left is a supervised English parser.}
\end{figure}

Figure~\ref{fig:tgt} offers a fine-grained look at which real and synthetic source languages $S$ succeeded best when $T=\text{English}$.  Each curve shows a different superstrate, with the $x$-axis ranging over substrates.  (The figure omits the hundreds of synthetic source languages that use two distinct superstrates, $\RV \neq \RN$.)  Real languages are shown as solid black dots, and are often beaten by synthetic languages.  For comparison, this graph also plots results that ``cheat'' by using English supervision.

The above graphs are evaluated on development sentences in development languages.  For our final results, Table~\ref{tb:finaltest}, we finally allow ourselves to try transferring to the UD test languages, and we evaluate on test sentences.  The comparison is similar to the comparison in the selection graph: do the synthetic treebanks add value?  We use our largest source pools, $n=10$ and $m=1210$.  With supervised selection, selecting the source language from the full pool of $m$ options (not just the $n$ real languages) tends to achieve significantly better UAS on the target language, often dramatically so.  On average, the UAS on the test languages increases by 2.3 percentage points, and this increase is statistically significant across these 17 data points.  Even with unsupervised selection, UAS still increases by 1.2 points on average, but this difference could be a chance effect.

\begin{table}[h!]
\setlength\tabcolsep{5pt}
\centering
\begin{small}
\begin{tabular}{l|cc|cc}
&\multicolumn{2}{c|}{unsupervised}&\multicolumn{2}{c}{(weakly) supervised}\\\hline 
target          &real   &+synthetic  &real    &+synthetic\\\hline\hline
el&60.07&\bf 65.72&\bf 65.87&\bf 66.98\\
he&\bf 63.39&60.65&62.86&\bf 64.28\\
la&46.79&\bf 51.82&56.62&\bf 59.06\\
hr&\bf 68.69&\bf 68.89&\bf 68.69&\bf 69.11\\
sv&74.96&74.96&74.96&74.96\\
hu&56.41&\bf 64.67&56.72&\bf 66.22\\
fa&53.41&\bf 58.37&53.41&\bf 60.18\\
fi\_ftb&50.90&\bf 55.36&53.03&\bf 55.86\\
cu&54.11&\bf 57.89&54.11&\bf 59.28\\
ga&53.55&\bf 59.38&57.72&\bf 64.72\\
sl&80.41&80.41&80.41&80.41\\
eu&47.12&\bf 48.97&45.35&\bf 52.90\\
ro&\bf 66.33&\bf 68.01&\bf 71.38&\bf 69.19\\
ja\_ktc&\bf 62.51&54.04&\bf 62.51&\bf 62.49\\
id&\bf 63.79&61.89&65.36&65.36\\
pl&\bf 75.69&\bf 74.63&\bf 75.69&73.05\\
ta&\bf 63.15&56.20&63.15&63.15\\\hline
Test Avg.&\bf 61.25&\bf 62.46&62.81&\bf 65.13\\\hline\hline
bg&\bf 79.80&74.52&79.80&79.80\\
nl&\bf 58.44&\bf 57.94&\bf 58.44&\bf 57.85\\
et&\bf 68.83&\bf 72.21&68.83&\bf 74.75\\
la\_proiel&48.50&\bf 49.66&48.50&\bf 49.66\\
da&71.65&71.65&\bf 71.65&\bf 70.79\\
en&\bf 63.37&61.37&63.37&\bf 65.43\\
grc&42.59&42.59&46.15&\bf 47.84\\
grc\_proiel&\bf 50.76&\bf 51.29&52.04&\bf 54.06\\
fi&51.28&\bf 55.21&\bf 54.46&\bf 55.21\\
got&54.98&\bf 57.57&54.98&\bf 58.66\\\hline
All Avg.&\bf 60.43&\bf 61.33&61.71&\bf 63.75\\ \hline\hline
\end{tabular}
\end{small}
\caption{\label{tb:finaltest}Our final comparison on the 17 test languages appears in the upper part of this table.  We ask whether single-source transfer to these 17 real target languages is improved by augmenting the source pool of 10 real languages with 1200 synthetic languages.  When different languages are selected in these two settings, we boldface the setting with higher test UAS, or both settings if they are not significantly different (paired permutation test by sentence, $p < 0.05$).  For completeness, we extend the table with the 10 development languages.  The ``Avg.'' lines report the average of 17 test or 27 test+dev languages.  The two supervised-selection averages are significantly different (paired permutation test by language, $p < 0.05$).}
\end{table}

The results above use gold POS tag sequences for $T$.  These may not be available if $T$ is a low-resource language; see Appendix~\ref{app:noisypos} for a further experiment.

\subsection{Discussion}

Many of the curves in Figures~\ref{fig:kite}--\ref{fig:selection} still seem to be increasing steadily at maximum $m$, which suggests that we would benefit from finding ways to generate even more synthetic languages.  Diversity of languages seems to be crucial, since adding new real languages improves performance much faster than remixing existing languages.  This suggests that we should explore making more extensive changes to the UD treebanks (see \S\ref{sec:futurework}).

Surprisingly, Figures~\ref{fig:kite}--\ref{fig:selection} show improvements even when $n=1$.  Evidently, self-permutation of a single language introduces some useful variety, perhaps by transporting specialized word orders (e.g., English still allows some limited V2 constructions) into contexts where the source language would not ordinarily allow them but the target language does.

Figure~\ref{fig:kite} shows why unsupervised selection is considerably worse on average than supervised selection.  Its 90th percentile is comparable, but at the 10th percentile---presumably representing experiments where no good sources are available---the unsupervised heuristic has more trouble at choosing among the mediocre options.  The supervised method can actually test these options using the true loss function.

Figure~\ref{fig:tgt} is interesting to inspect. 
English is essentially a Germanic language with French influence due to the Norman conquest, so it is reassuring that German and French substrates can each be improved by using the other as a superstrate.  We  also see that Arabic and Hindi are the worst source languages for English, but that Hindi[Arabic/V] is considerably better.  This is because Hindi is reasonably similar to English once we correct its SOV word order to SVO (via almost any superstrate).

\section{Conclusions and Future Work}\label{sec:futurework}

This paper is the first release of a novel resource, the \ourData treebank collection, that may unlock a wide variety of research opportunities (discussed in \S\ref{sec:intro}). Our empirical studies show that the synthetic languages in this collection remain somewhat natural while improving the diversity of the collection.  As a simplistic but illustrative use of the resource, we carefully evaluated its impact on the naive technique of single-source transfer parsing.  We found that performance could consistently be improved by adding synthetic languages to the pool of sources, assuming gold POS tags.

There are several non-trivial opportunities for improving and extending our treebank collection in future releases.  

1.~Our current method is fairly conservative, only synthesizing languages with word orders already attested in our small collection of real languages.  This does not increase the diversity of the pool as much as when we add new real languages.  Thus, we are particularly interested in generating a wider range of synthetic languages.  We could condition reorderings on the surrounding tree structure, as noted in \S\ref{sec:proscons}.  We could choose reordering parameters $\vtheta_X$ more adventurously than by drawing them from a single known superstrate language.  We could go beyond reordering, to systematically choose what function words (determiners, prepositions, particles), function morphemes, or punctuation symbols
\footnote{Our current handling of punctuation produces unnatural results, and not merely because we treat all tokens with tag \textsc{punct} as interchangeable.  Proper handling of punctuation and capitalization would require more than just reordering.  For example, ``\texttt{Jane loves her dog, Lexie.}'' should reorder into ``\texttt{Her dog, Lexie, Jane loves.}'', which has an extra comma and an extra capital.  Accomplishing this would require first recovering a richer tree for the original sentence, in which the appositive \texttt{Lexie} is bracketed by a {\em pair} of commas and the name \texttt{Jane} is {\em doubly} capitalized.  These extra tokens were not apparent in the original sentence's surface form because the final comma was absorbed into the adjacent period, and the start-of-sentence capitalization was absorbed into the intrinsic capitalization of \texttt{Jane} \cite{nunberg-1990}.  The tokenization provided by the UD treebanks unfortunately does not attempt to undo these orthographic processes, even though it undoes some morphological processes such as contraction.}
 should appear in the synthetic tree, or to otherwise alter the structure of the tree \cite{dorr-1993}.  These options may produce implausible languages.  To mitigate this, we could filter or reweight our sample of synthetic languages---via rejection sampling or importance sampling---so that they are distributed more like real languages, as measured by their parsabilities, dependency lengths, and estimated WALS features \cite{wals}.

2.~Currently, our reordering method only generates {\em projective} dependency trees.  We should extend it to allow non-projective trees as well---for example, by pseudo-projectivizing the substrate treebank \cite{nivre-nilsson-2005} and then deprojectivizing it after reordering.

3.~The treebanks of real languages can typically be augmented with larger unannotated corpora in those languages \cite{W2C-2011}, which can be used to train word embeddings and language models, and can also be used for self-training and bootstrapping methods.  We plan to release comparable unannotated corpora for our synthetic languages, by automatically parsing and permuting the unnanotated corpora of their substrate languages.

4.~At present, all languages derived from an English substrate use the English vocabulary.  In the future, we plan to encipher that vocabulary separately for each synthetic language, perhaps choosing a cipher so that the result loosely conforms to the realistic phonotactics and/or orthography of some superstrate language.  This would let multilingual methods exploit lexical features without danger of overfitting to specific lexical items that appear in many synthetic training languages.  Alphabetic ciphers can preserve features of words that are potentially informative for linguistic structure discovery: their cooccurrence statistics, their length and phonological shape, and the sharing of substrings among morphologically related words.

5.~Finally, we note that this paper has focused on generating a broadly reusable collection of synthetic treebanks.  For some applications (including single-source transfer), one might wish to tailor a synthetic language on demand, e.g., starting with one of our treebanks but modifying it further to more closely match the surface statistics of a given target language \cite{dorr-et-al-2002}.  In our setup, this would involve actively searching the space of reordering parameters, using algorithms such as gradient ascent or simulated annealing.

We conclude by revisiting our opening point.  Unsupervised discovery of linguistic structure is difficult.  We often do not know quite what function to maximize, or how to globally maximize it.  If we could make labeled languages as plentiful as labeled images, then we could treat linguistic structure discovery as a problem of {\em supervised} prediction---one that need not succeed on all formal languages, but  which should generalize at least to the domain of {\em possible} human languages.

\appendix

\section{Constructing the Kite Graph}\label{app:kite}

The mean lines in the ``kite graph'' (Figure~\ref{fig:kite}) are actually obtained by averaging 10,000 graphs.  Each of these graphs is ``smooth'' because it incrementally adds new languages as $n$ or $m$ increases.  Pseudocode to generate one such graph is given as Algorithm~\ref{alg:exp}; all random choices are made uniformly.

\addtolength{\textfloatsep}{-10pt}
\begin{algorithm}
    \caption{\label{alg:exp}Data collection for one graph}
\begin{small}
    \begin{algorithmic}[1]
      \INPUT Sets ${\cal T}$ (target languages), ${\cal S}$ (real source languages), ${\cal S}'$ (synthetic source languages)
      \OUTPUT Sets of data points $D_{\text{sup}}$, $D_{\text{unsup}}$
      \Procedure{CollectData}{}
        \State $D \gets \varnothing$ 
        \State Sample a target language $T$ from $\cal T$
        \State $L \gets \text{random.shuffle}(\mc{S}-\{T\})$ 
        \State $L'\gets \text{random.shuffle}(\mc{S'})$
        \For{$n = 1$ \textbf{to} $|L|$}
        \State $L'' \gets {}$\parbox[t]{2in}{a filtered version of $L'$ that excludes languages with substrates or superstrates outside $\{L_1,\ldots,L_n\}$}
        \For{$n' = 1$ \textbf{to} $|L''|$}
        \State ${\cal P} \gets \{L_1, \ldots, L_n, L''_1,\ldots,L''_{n'}\}$
        \State $m \gets |{\cal P}|$
        \State $D_{\text{sup}} \gets {}$\parbox[t]{1in}{$D_{\text{sup}} \cup \\ \{(n, m, \textsc{uas}_{\text{sup}}({\cal P},T))\}$}
        \State $D_{\text{unsup}} \gets {}$\parbox[t]{1in}{$D_{\text{unsup}} \cup \\ \{(n, m, \textsc{uas}_{\text{unsup}}({\cal P},T))\}$}
        \EndFor
        \EndFor
        \State \Return $(D_{\text{sup}}$, $D_{\text{unsup}})$
    \EndProcedure
  \end{algorithmic}
\end{small}
\end{algorithm}

\section{Experiment with Noisy Tags}\label{app:noisypos}

Table \ref{tb:error_finaltest} repeats the single-source transfer experiment using noisy automatic POS tags for $T$ for both parser input and unsupervised selection.  We obtained the tags using RDRPOSTagger \cite{nguyen-EtAl:2014:Demos} trained on just 100 gold-tagged sentences (the same set used for supervised selection).  
The low tagging accuracy does considerably degrade UAS and muddies the usefulness of the synthetic sources.

\begin{table}[!h]
\setlength\tabcolsep{3pt}
\begin{tabular}{l|c|cc|cc}
&tag&\multicolumn{2}{c|}{unsupervised}&\multicolumn{2}{l}{(weakly) superv.}\\\hline 
target&          &real   &+synth  &real    &+synth\\\hline\hline
bg&78.33&53.24&\bf 55.08&53.24&53.24\\
nl&71.70&\bf 39.40&\bf 38.99&\bf 42.42&\bf 42.75\\
et&72.88&45.19&\bf 54.81&\bf 56.07&\bf 55.09\\
la\_proiel&71.83&37.25&\bf 38.26&37.25&\bf 38.10\\
da&78.04&\bf 47.98&43.40&\bf 47.98&45.89\\
en&77.33&\bf 48.29&44.40&\bf 48.29&\bf 48.15\\
grc&68.80&32.15&32.15&\bf 33.52&\bf 34.36\\
grc\_proiel&72.93&\bf 42.46&41.39&43.49&\bf 44.19\\
fi&65.65&\bf 29.59&\bf 28.81&\bf 36.85&\bf 36.90\\
got&76.66&\bf 44.77&\bf 44.05&44.77&\bf 46.83\\\hline
Avg.&73.42&\bf 42.03&\bf 42.13&\bf 44.39&\bf 44.55\\ \hline\hline
\end{tabular}
\caption{\label{tb:error_finaltest}Tagging accuracy on the 10 dev languages, and UAS of the selected source parser with these noisy target-language tag sequences.  The results are formatted as in Table~\ref{tb:finaltest}, but here all results are on dev sentences.}
\end{table}

\newpage
\paragraph{Acknowledgements} This work was funded by the U.S. National Science Foundation under Grant No. 1423276.  Our data release is derived from the Universal Dependencies project, whose many selfless contributors have our gratitude.  We would also like to thank Matt Gormley and Sharon Li for early discussions and code prototypes, Mohammad Sadegh Rasooli for guidance on working with the Yara parser, and Jiang Guo, Tim Vieira, Adam Teichert, and Nathaniel Filardo for additional useful discussion. Finally, we thank TACL editors Joakim Nivre and Lillian Lee and the anonymous reviewers for several suggestions that improved the paper.

\small
\bibliography{galactic}{}
\bibliographystyle{plainnaturl}

\end{document}